\documentclass{article}
\pdfoutput=1
\usepackage{spconf,amsmath,graphicx}

\usepackage{epsfig}
\usepackage{amssymb}
\usepackage{array}

\title{ERNET Family: Hardware-Oriented CNN Models for Computational Imaging Using Block-Based Inference}
\name{Chao-Tsung Huang\thanks{This work was supported by the Ministry of Science and Technology, Taiwan, R.O.C. under Grant no. MOST 108-2218-E-007-020.}}
\address{National Tsing Hua University, Taiwan, R.O.C.}
\begin{document}
\ninept
\maketitle
\begin{abstract}
Convolutional neural networks (CNNs) demand huge DRAM bandwidth for computational imaging tasks, and block-based processing has recently been applied to greatly reduce the bandwidth.
However, the induced additional computation for feature recomputing or the large SRAM for feature reusing will degrade the performance or even forbid the usage of state-of-the-art models.
In this paper, we address these issues by considering the overheads and hardware constraints in advance when constructing CNNs.
We investigate a novel model family---ERNet---which includes temporary layer expansion as another means for increasing model capacity.
We analyze three ERNet variants in terms of hardware requirement and introduce a hardware-aware model optimization procedure.
Evaluations on Full HD and 4K UHD applications will be given to show the effectiveness in terms of image quality, pixel throughput, and SRAM usage.
The results also show that, for block-based inference, ERNet can outperform the state-of-the-art FFDNet and EDSR-baseline models for image denoising and super-resolution respectively.

\end{abstract}
%
\section{Introduction}
\label{sec:intro}

Convolutional neural networks (CNNs) have recently demonstrated superior quality for not only computer vision but also computational imaging applications.
Their success in the computer vision field, such as image recognition \cite{vggnet_2015,resnet_2016} and object detection \cite{detection_review_2019}, has led to the rising wave of deep learning.
On the other hand, CNNs have also taken the image quality of computational imaging into a new level, e.g.\ image denoising \cite{DnCNN_2017,FFDNet_2018}, super-resolution (SR) \cite{VDSR_2016,SRResNet_2017,EDSR_2017,WDSR_2018}, style transfer \cite{st_sr_2016,cyclegan_2017}, and algorithm mimicking \cite{ipcopy_2017}.
Therefore, they have great potential to bring an image pipeline revolution on cameras and displays for our daily use if their edge inference with state-of-the-art quality is possible. 

The most effective way of enabling high-quality inference at resource-limited edge devices is to construct hardware-friendly CNN models.
Most of previous works focused on computer vision models and aimed to reduce complexity based on model sparsity.
For example, MobileNet \cite{mobilenet_2017} applies depth-wise convolution to perform low-rank computation, and SqueezeNet \cite{squeezenet_2017} temporarily reduces model width and then expands back for residual connections.
Furthermore, MobileNetV2 \cite{mobilenetv2_2018} moves the connections to thinner layers to reduce storage and results in an expansion-reduction structure for its building block.
On the other hand, the complexity can also be reduced by pruning small weights \cite{deepcompression_2016}.

However, most of these complexity-reducing techniques do not apply to computational imaging models because the sparsity assumption could become invalid.
For instance, weight pruning and depth-wise convolution were found to cause serious quality degradation for denoising and SR \cite{eCNN_2019}.
Moreover, the huge DRAM bandwidth demanded by high-end applications, which is a major bottleneck for edge inference, was not considered at all.
Recently, a network called WDSR \cite{WDSR_2018} was proposed to restructure SR models using also feature expansion/reduction and low-rank convolution for better tradeoffs between quality and complexity; nevertheless, the bandwidth issue was still not discussed.

\begin{figure}[t]
\begin{minipage}[b]{.32\linewidth}
  \centering
  \centerline{\includegraphics[width=2.6cm]{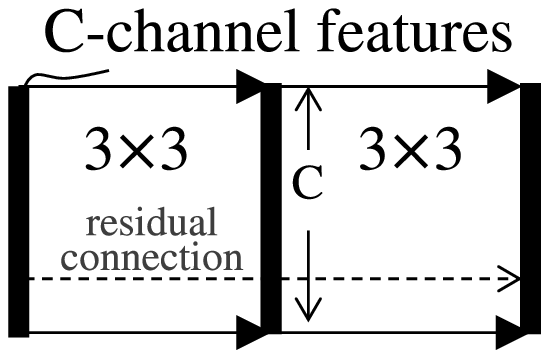}}
  \centerline{(a) ResNet \cite{resnet_2016}}\medskip
\end{minipage}
\hfill
\begin{minipage}[b]{0.32\linewidth}
  \centering
  \centerline{\includegraphics[width=2.6cm]{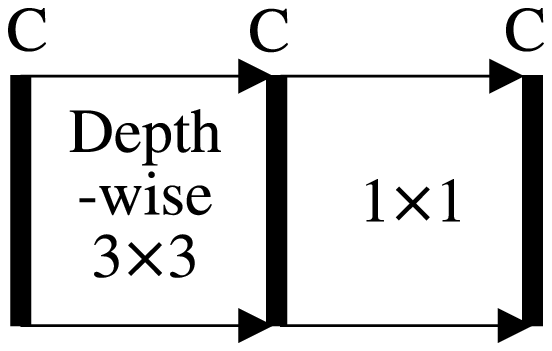}}
  \centerline{(b) MobileNet \cite{mobilenet_2017}}\medskip
\end{minipage}
\hfill
\begin{minipage}[b]{0.32\linewidth}
  \centering
  \centerline{\includegraphics[width=2.6cm]{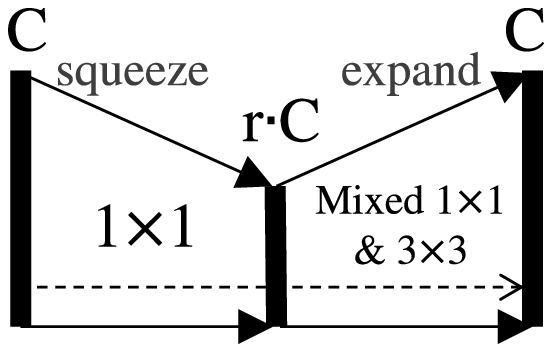}}
  \centerline{(c) SqueezeNet \cite{squeezenet_2017}}\medskip
\end{minipage}
\vfill
\begin{minipage}[b]{.32\linewidth}
  \centering
  \centerline{\includegraphics[width=2.6cm]{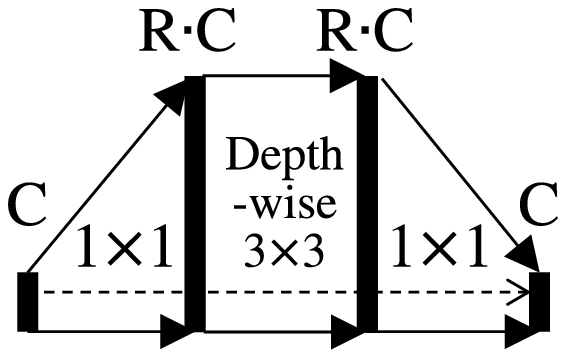}}
  \centerline{(d) MobileNetV2 \cite{mobilenetv2_2018}}\medskip
\end{minipage}
\hfill
\begin{minipage}[b]{0.32\linewidth}
  \centering
  \centerline{\includegraphics[width=2.6cm]{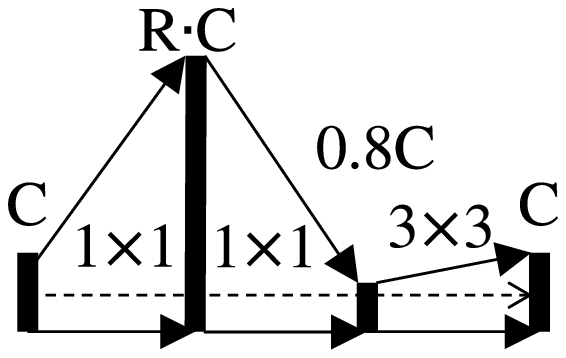}}
  \centerline{(e) WDSR-B \cite{WDSR_2018}}\medskip
\end{minipage}
\hfill
\begin{minipage}[b]{0.32\linewidth}
  \centering
  \centerline{\includegraphics[width=2.6cm]{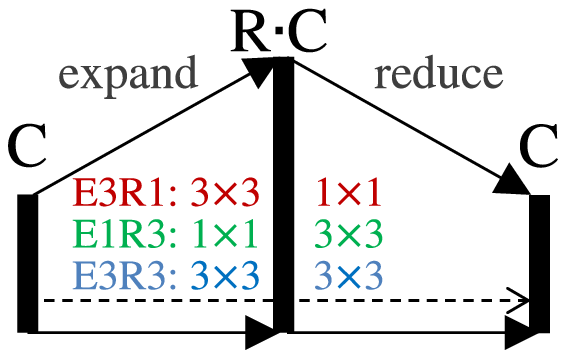}}
  \centerline{(f) Proposed ERNet}\medskip
\end{minipage}
\vspace*{-0.3cm}
\caption{Building blocks of different networks with their model widths (feature channel numbers) and convolution filters.
(a) Residual blocks with the same width and only 3$\times$3 filters.
(b) Depth-wise convolution.
(c) Feature squeezing and then expanding.
(d-e) Feature expanding and then reducing both with 1$\times$1 filters.
(f) Three variants of the proposed expansion-reduction structure. 
}
\label{fig:fig_channel}
\end{figure}

For computational imaging models, the conventional layer-by-layer inference flow demands huge DRAM bandwidth for feature maps.
And block-based processing can eliminate this traffic by computing convolution layers altogether for each small image block.
For example, the layer fusion in \cite{fusedlayer_2016} proposes a pyramid inference flow and suggests to reuse the overlapped features between moving blocks.
In contrast, the truncated-pyramid inference flow in \cite{eCNN_2019} proposes to recompute these features for saving SRAM area.
But the recomputing overhead is increased quickly with model depth and thus could forbid the usage of deep networks.

In this paper, we aim to investigate efficient model construction for high-resolution computational imaging tasks.
In particular, we focus on the models designed for block-based processing flows to enable bandwidth-efficient edge inference.
Unlike the previous works which passively slim or restructure existing networks, we actively construct hardware-efficient models by using the expansion-reduction structure in Fig. \ref{fig:fig_channel}(f) and considering the expansion ratio as a major model hyperparameter.
The main contribution of this paper is to extend the basic idea of ERNet in our previous work \cite{eCNN_2019} (E3R1 variant for only feature recomputing) to a comprehensive study which further includes two more variants, the feature reusing scheme, SRAM requirement analysis, and extensive evaluations for denoising and SR applications.

The rest of this paper is organized as follows.
In Section \ref{sec:blk_flow}, two block-based inference flows using feature recomputing and reusing, respectively, are introduced.
In Section \ref{sec:ernet}, we discuss the ERNet family in terms of model structure, SRAM requirement, and model optimization.
Extensive evaluations on Full-HD and 4K-UHD use cases will be presented in Section \ref{sec:eval} to show the advantages over the state-of-the-art FFDNet \cite{FFDNet_2018} and EDSR-baseline \cite{EDSR_2017} for denoising and SR respectively.
Finally, concluding remarks are given in Section \ref{sec:conclusion}.

\section{Block-based inference flows}
\label{sec:blk_flow}

The conventional inference flow is to compute features layer-by-layer for one whole image.
It reuses parameters efficiently but consumes huge DRAM bandwidth for storing and reading back the features in the internal layers.
For example, such bandwidth for supporting FFDNet (12-layer, 96-channel) at 4K UHD 30fps is up to
\begin{align}
3840/2 \cdot 2160/2 \cdot 96\mbox{Bytes} \cdot (12-1) \cdot 30\mbox{Hz} \cdot 2 = 131 \mbox{GB/s}, \label{equ:bandwidth}
\end{align}
where 8-bit features are assumed.
To eliminate such bandwidth, we consider block-based inference which partitions one image into several blocks and performs layer-by-layer inference for each block.
Then the features can be stored in on-chip buffers, instead of DRAM.

However, the block-based inference induces two overheads.
The first one is on-chip block buffers to store intermediate features.
It will affect the efficiency of parameter reuse since each block needs to reload the whole model again; therefore, a small block size could deteriorate computing performance.
The second one is the need to handle the overlapped features between block boundaries since the receptive field is larger than one pixel.
To simplify the discussion, we will consider the case in which we have a sufficiently large block size and focus on the feature overlapping issue in the following.

\begin{figure}[t]
\begin{minipage}[b]{.56\linewidth}
  \centering
  \centerline{\includegraphics[width=4.48cm]{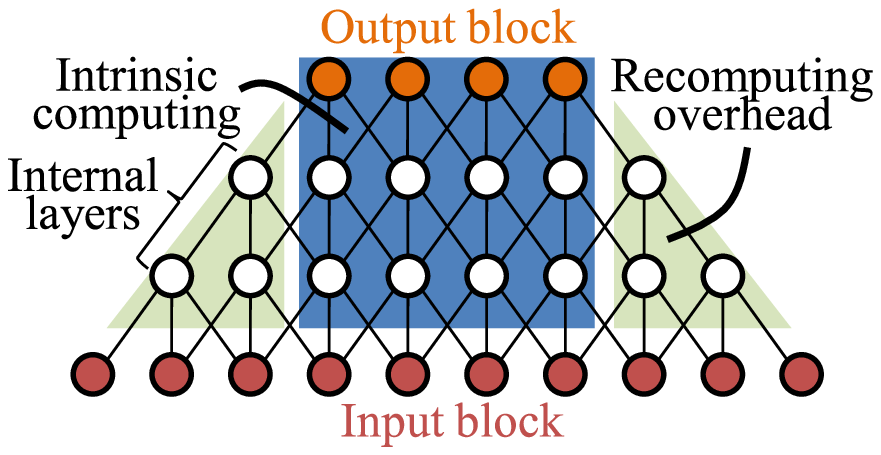}}
  \centerline{(a) Feature recomputing}\medskip
\end{minipage}
\hfill
\begin{minipage}[b]{0.43\linewidth}
  \centering
  \centerline{\includegraphics[width=3.44cm]{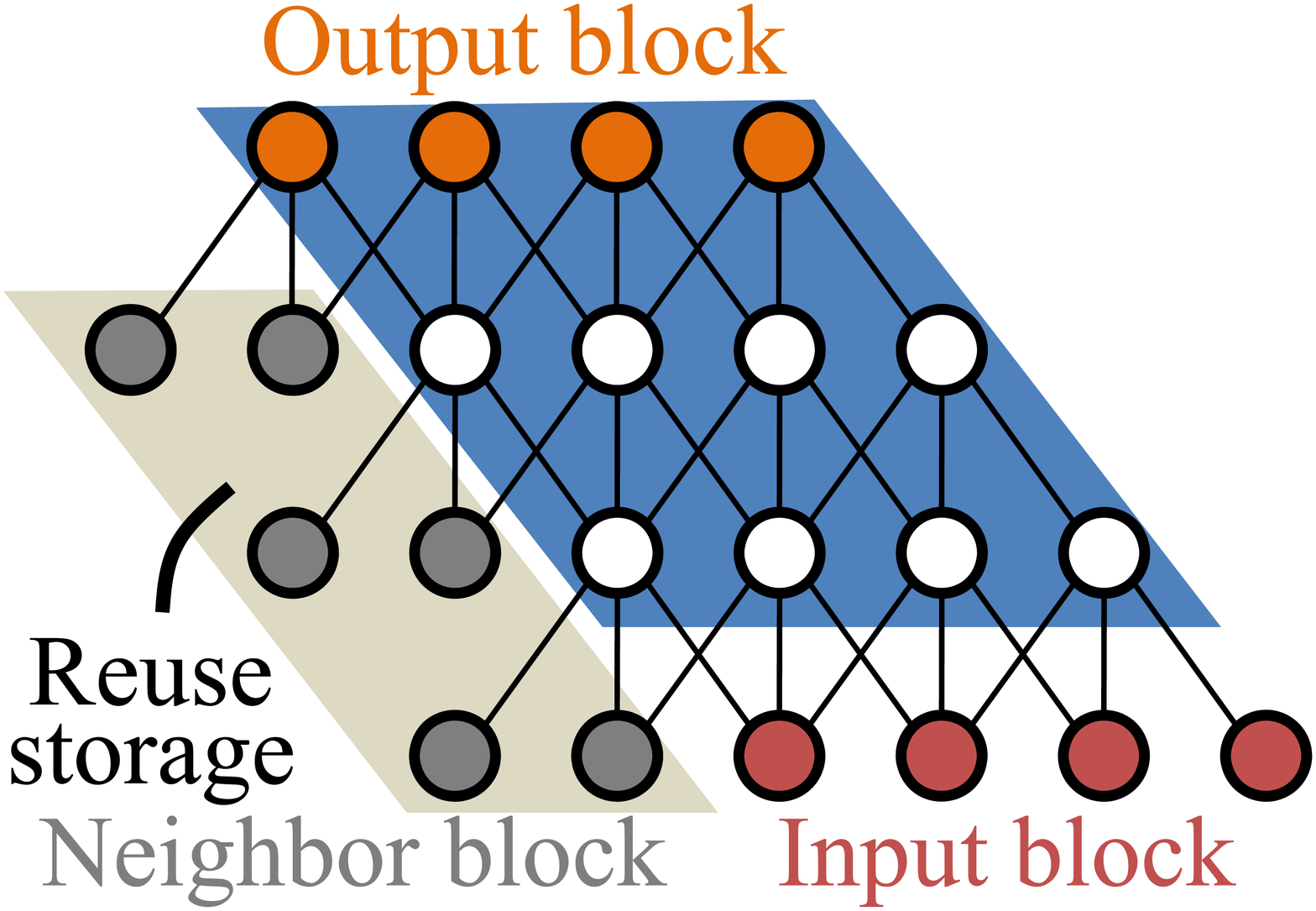}}
  \centerline{(b) Feature reusing}\medskip
\end{minipage}
\vspace*{-0.6cm}
\caption{Cross-sectional views for two block-based flows with 3$\times$3 filters. Each circle represents an input/output pixel or a feature. The inference in one block is performed from the input block (red) at the bottom to the output block (orange) at the top.}
\label{fig:fig_blk_inf}
\end{figure}

Two different approaches can be applied to handle the overlapped features.
One is to recompute them for each block as \cite{eCNN_2019}, and the other one is to reuse them with additional storage similarly to the layer fusion \cite{fusedlayer_2016}.
To illustrate the corresponding inference flows in one block, we show their 1-D cross-sectional views in Fig. \ref{fig:fig_blk_inf}.
The feature recomputing results in a truncated-pyramid flow for which the processed block will get smaller when going to deeper layers. In contrast, the feature reusing forms an oblique-cuboid flow which has the same input and output block size.

Now we can analyze their overheads for evaluating design tradeoffs.
The recomputing approach demands additional computation (green part in Fig. \ref{fig:fig_blk_inf}(a)), and the amount will increase quickly as model depth goes deeper.
For example, consider a plain network with only 3$\times$3 filters.
The ratio of the recomputing overhead to the original complexity is $\frac{2}{3} \frac{\beta (3- 4 \beta)}{(1-2 \beta)^2}$ where $\beta=\frac{D}{S}$ for model depth $D$ and block width $S$.
With a 128$\times$128 block, the ratio is only 0.5 for a 20-layer plain network but it will go up to 2.6 for a 40-layer one.
Therefore, deep networks could be unfavorable.

On the other hand, the reusing approach avoids additional computing by storing the already-computed features for neighbor blocks.
Assume we move the blocks from left to right for a plain network.
For reusing in the vertical direction, we need to store two horizontal lines of features in width $W$ for each of input or internal layers.
Similarly, two another vertical stripes in block height $S$ are required for the horizontal reuse.
For a plain $D$-layer $C$-channel network with an image channel number $C_{in}$, the size of the line buffers will be as high as $2(W+S)(C_{in}+(D-1)C)$.
For example, FFDNet requires up to 4.0MB of SRAM as the line buffers for 4K UHD resolution.

\begin{figure}[t]
\begin{minipage}[b]{.48\linewidth}
  \centering
  \centerline{\includegraphics[width=4.0cm]{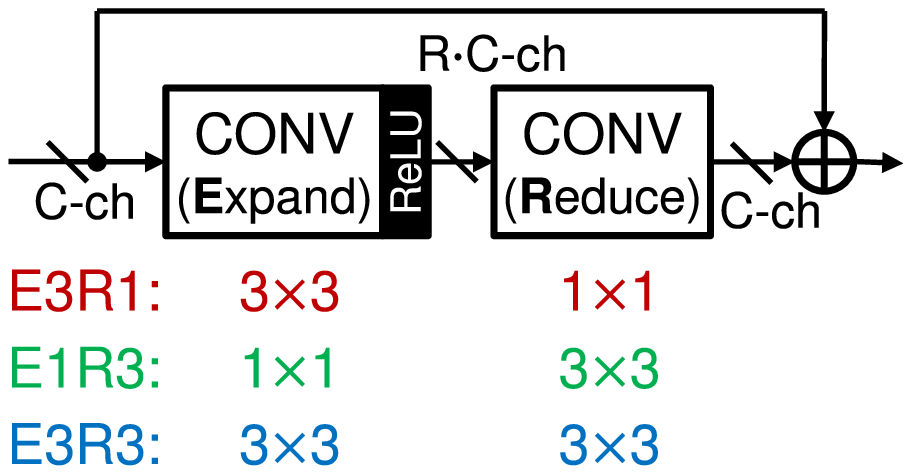}}
  \centerline{(a) ERModule}\medskip
\end{minipage}
\hfill
\begin{minipage}[b]{0.48\linewidth}
  \centering
  \centerline{\includegraphics[width=3.9cm]{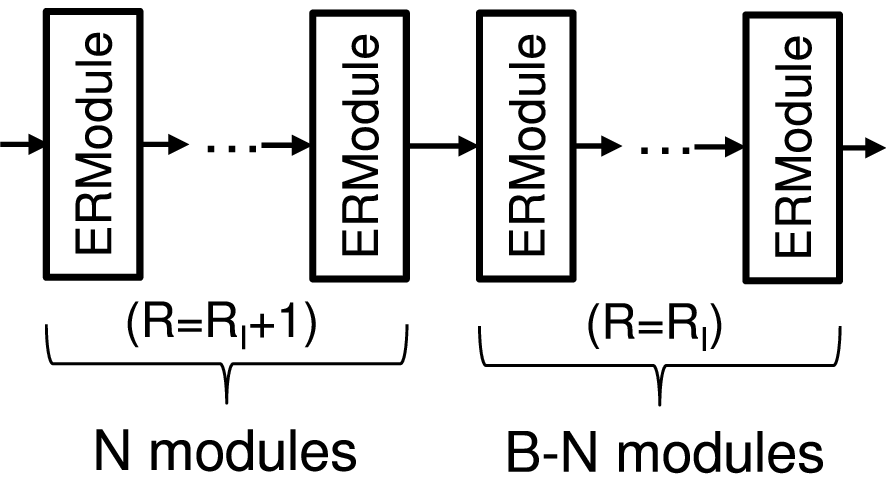}}
  \centerline{(b) Connected ERModules}\medskip
\end{minipage}
\vfill
\begin{minipage}[b]{.48\linewidth}
  \centering
  \centerline{\includegraphics[width=4.0cm]{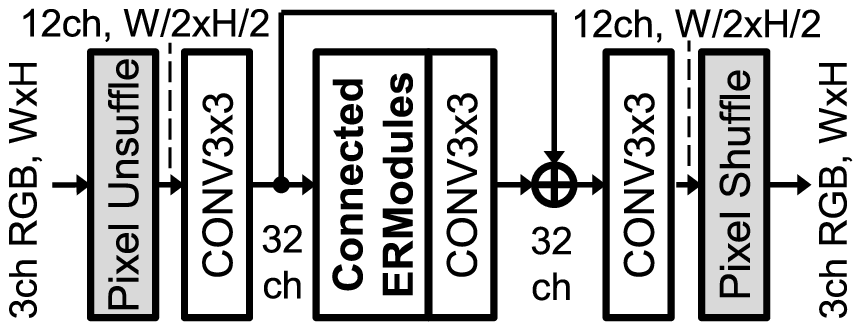}}
  \centerline{(c) DnERNet-12ch}\medskip
\end{minipage}
\hfill
\begin{minipage}[b]{0.48\linewidth}
  \centering
  \centerline{\includegraphics[width=4.0cm]{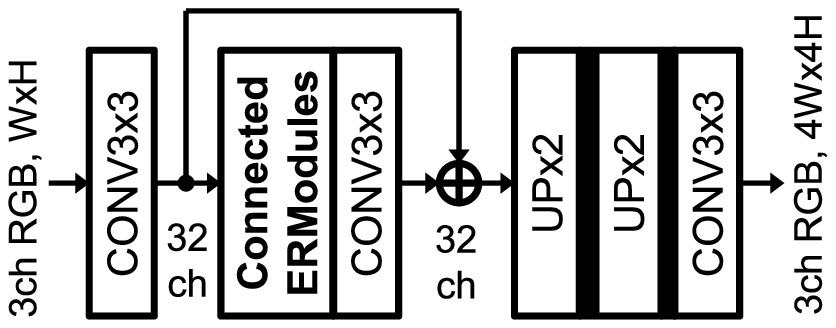}}
  \centerline{(d) SR4ERNet}\medskip
\end{minipage}
\vspace*{-0.3cm}
\caption{ERNet family. (a) Three variants of building blocks (ERModules): E3R1, E1R3, and E3R3. (b) Connecting several ERModules for a fractional expansion ratio $R_{I}\frac{N}{B}$. (c) Exemplar denoising network using ERNet. (d) Exemplar SR$\times$4 network using ERNet.}
\label{fig:fig_model}
\end{figure}

\section{ERN\lowercase{et} family}
\label{sec:ernet}

In the following, we will propose a novel CNN family---ERNet---to construct hardware-efficient models which are aware of the overheads resulted from block-based inference.
We will first introduce the model structures and their SRAM requirement and then present hardware-oriented model optimization procedures.

\subsection{Model structure}
\label{ssec:structure}

Suppose we want to build a high-quality network and start from a small one which has few layers and few feature channels.
The conventional approach to add model capacity is to increase model depth $D$ and/or model width $C$; however, it could cause significant overheads for block-based inference.
For example, the block buffer size is proportional to $C$ and the line buffer size for feature reusing is nearly proportional to $D\cdot C$.
Also, the computation overhead for feature recomputing is increased very quickly for a large depth $D$.

To overcome this difficulty, we propose to use temporary layer expansion as an additional means for model construction.
Instead of simply adding depth or width, we can pump capacity into the network by temporarily expanding the channels as shown in Fig. \ref{fig:fig_channel}(f).
Since the channel reduction is performed immediately after the expansion, the effective model width is not enlarged at all.
In other words, we now have the expansion ratio as an additional model hyperparameter besides depth $D$ and width $C$.

With this idea, we devise three variants of build blocks named ERModules as shown in Fig. \ref{fig:fig_model}(a).
They contain at least one 3$\times$3 filter to increase receptive field and use another 1$\times$1 or 3$\times$3 filter for the expansion-reduction purpose.
For increasing model flexibility, we may use fractional expansion ratios.
However, this could cause low hardware utilization for highly-parallel acceleration.
Instead, we consider only integer expansion ratios in every ERModule and construct a larger building block by connecting $B$ of them as shown in Fig. \ref{fig:fig_model}(b).
Now we can have an equivalent fractional expansion ratio $R_E=R_I\frac{N}{B}$ by setting the expansion ratios of the first $N$ ERModules as $R_I+1$ and those of the rest as $R_I$.

Based on the connected ERModules, we construct two ERNets for denoising and SR$\times$4, respectively, in Fig. \ref{fig:fig_model}(c-d).
The DnERNet-12ch uses the same downsampling strategy in FFDNet \cite{FFDNet_2018} with an additional skip connection, and the SR4ERNet replaces the ResBlocks in EDSR-baseline \cite{EDSR_2017} by ERMdoules.
We use only 32-ch features for the input and output of ERModules to save on-chip buffers.
Finally, we have two model hyperparameters to build networks: $B$ for increasing depth and $R_E$ for pumping complexity.

\subsection{SRAM requirement}
\label{ssec:er_require}

The computing overheads for the feature recomputing flow are mainly related to the number of 3$\times$3 layers.
However, the required line buffer sizes for the feature reusing flow are quite different for the three ERModule variants.
Their SRAM requirement is summarized in Table \ref{tab:tab_sram_req}.
For comparison, we also include the conventional building block CONV3$\times$3 ($R^{0.5}C$-ch to $R^{0.5}C$-ch 3$\times$3 filter).
Note that the E3R1 and E1R3 variants needs 11\% of more computation for one equivalent 3$\times$3 layer due to the reduction and expansion 1$\times$1 filters respectively.

ERModules can use smaller block buffers than the CONV3$\times$3 building block as expected.
For the feature reusing flow, the required line buffer size is determined be the input channel number of each 3$\times$3 filter.
For example, the 3$\times$3 filter of E3R1 has only $C$-ch input features while that of E1R3 works on expanded $RC$-ch features; therefore, E1R3 will require $R\times$ of line buffers.
For fair comparison, we consider the normalized line buffer sizes per 3$\times$3 layer.
Then the four building blocks in Table \ref{tab:tab_sram_req} can be ranked in order of the normalized sizes: E3R1, CONV3$\times$3, E3R3, and E1R3.
And their corresponding ratios are $1:R^{0.5}:\frac{1+R}{2}:R$.
As a result, we will prefer to use E3R1 for the feature reusing flow.
Incidentally, MobileNetV2 \cite{mobilenetv2_2018} will require the same large line buffers as E1R3 since it applies depth-wise 3$\times$3 filters on expanded channels.

\begin{table}[t]
\caption{SRAM requirement for one ERModule.}
\begin{minipage}[b]{1.0\linewidth}
  \centering
  \centerline{\includegraphics[width=8.4cm]{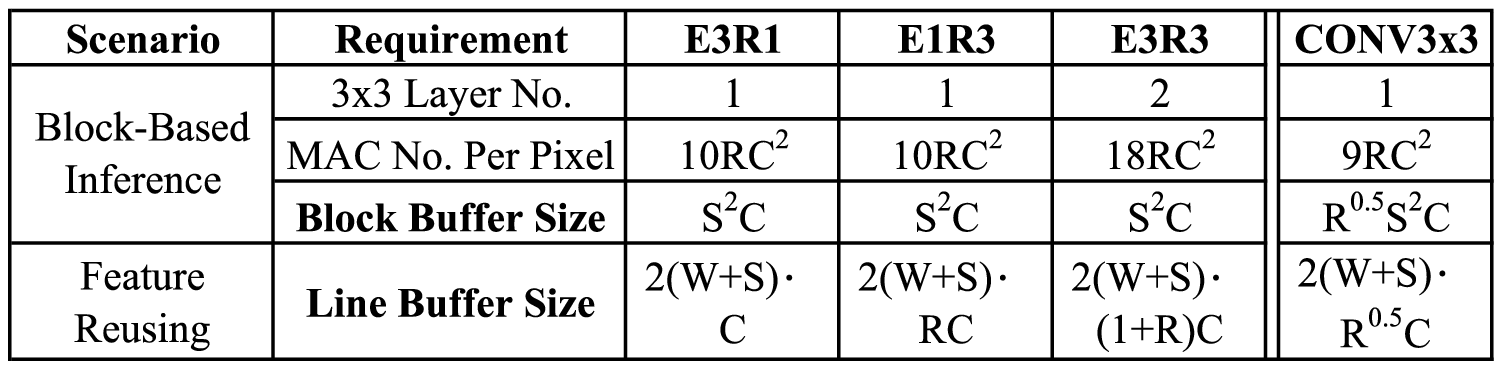}}
\end{minipage}
\vspace*{-0.75cm}
\label{tab:tab_sram_req}
\end{table}

\begin{table}[t]
\caption{Performance targets and computing capability.}
\begin{minipage}[b]{1.0\linewidth}
  \centering
  \centerline{\includegraphics[width=8.4cm]{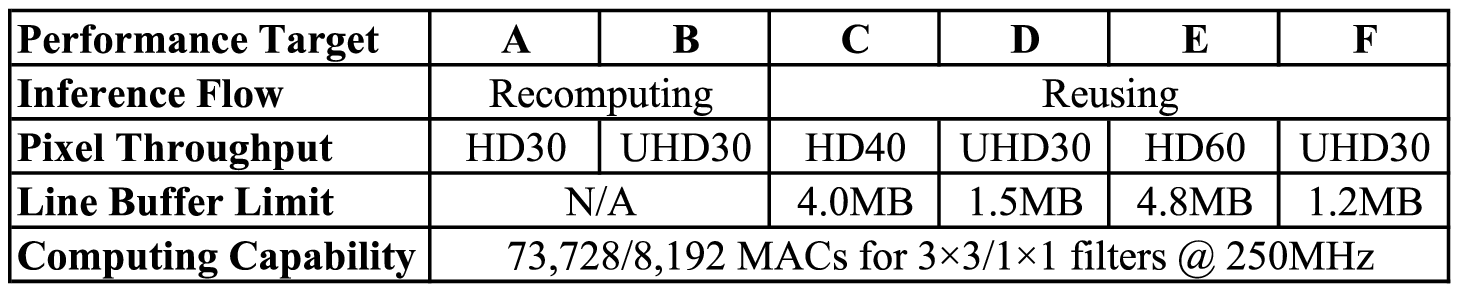}}
\end{minipage}
\vspace*{-0.75cm}
\label{tab:tab_hw_setting}
\end{table}

\subsection{Model optimization}
\label{ssec:model_opt}

We have two hyperparameters, $B$ and $R_E$, to build ERNets, and they can be used to optimize the models for given hardware constraints.
In particular, we can find a maximum expansion ratio $R_E$ for every considered ERModule number $B$ such that the total computation complexity is smaller than or equal to given computing capability.
Thus we can perform model scanning in terms of $B$, or the corresponding depth $D$, and pick the best model using validation quality.
After that, we can further polish the model with a heavier training setting to improve quality.

\begin{figure}[t]
\begin{minipage}[b]{.48\linewidth}
  \centering
  \centerline{\includegraphics[width=4.0cm]{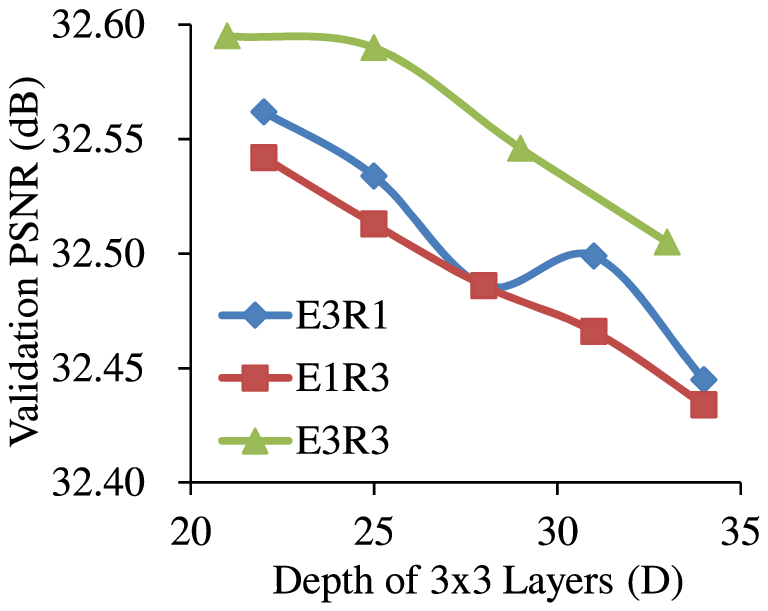}}
  \centerline{(a) HD30 (Target A)}\medskip
\end{minipage}
\hfill
\begin{minipage}[b]{0.48\linewidth}
  \centering
  \centerline{\includegraphics[width=4.0cm]{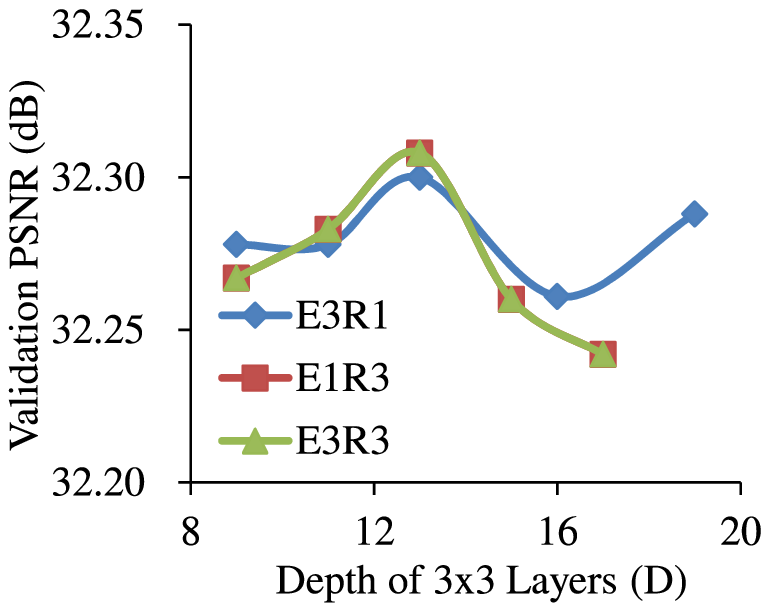}}
  \centerline{(b) UHD30 (Target B)}\medskip
\end{minipage}
\vspace*{-0.5cm}
\caption{DnERNet-12ch scanning for the feature recomputing flow. The performance targets are (a) Full HD 30fps (HD30) and (b) 4K UHD 30fps (UHD30). Validation with ten DIV2K images \cite{div2k}.}
\label{fig:fig_scan_FC}
\end{figure}

\begin{figure}[t]
\begin{minipage}[b]{.48\linewidth}
  \centering
  \centerline{\includegraphics[width=4.0cm]{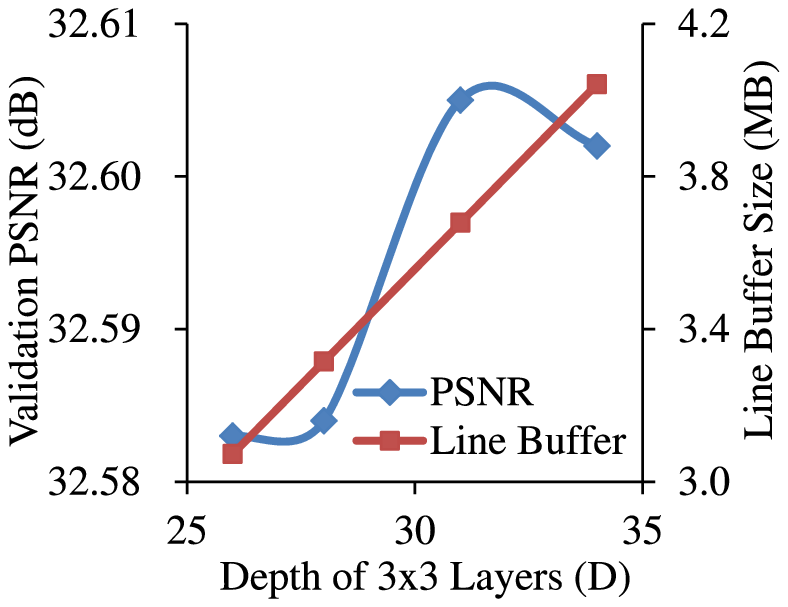}}
  \centerline{(a) Denoising, HD40 (Target C)}\medskip
\end{minipage}
\hfill
\begin{minipage}[b]{0.48\linewidth}
  \centering
  \centerline{\includegraphics[width=4.0cm]{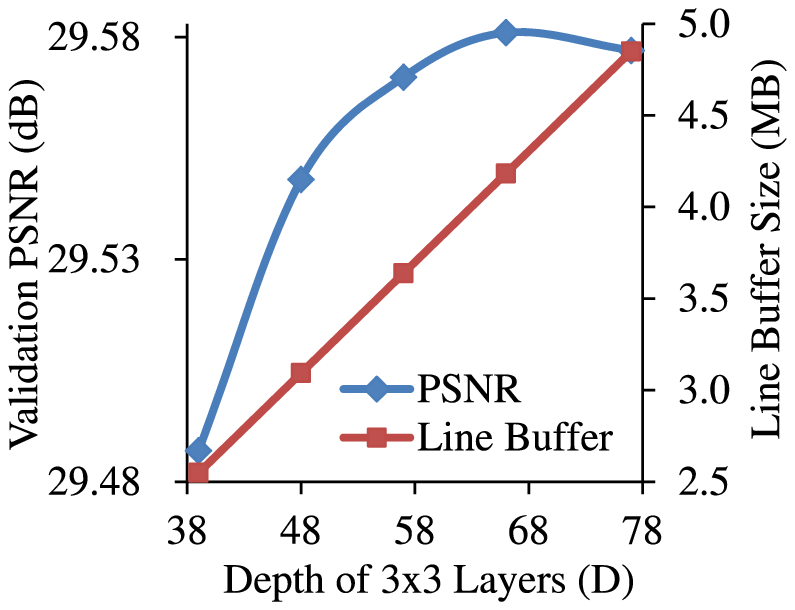}}
  \centerline{(b) SR$\times$4, HD60 (Target E)}\medskip
\end{minipage}
\vspace*{-0.5cm}
\caption{ERNet-E3R1 scanning for the feature reusing flow. The performance targets are (a) denoising at HD40 with $\leq$4.0MB of line buffers (LB4.0) and (b) SR$\times$4 at HD60 with LB4.8.}
\vspace*{-0.3cm}
\label{fig:fig_scan_FU}
\end{figure}

For feature recomputing, we show two examples in Fig. \ref{fig:fig_scan_FC} for DnERNet-12ch scanning to support Full HD and 4K UHD applications (performance targets A and B in Table \ref{tab:tab_hw_setting}).
The three variants all follow a similar trend: deeper networks are not always preferred any more due to their higher recomputing overheads.
Similarly, we show two examples for feature reusing in Fig. \ref{fig:fig_scan_FU} to support Full-HD denoising and SR tasks (targets C and E in Table \ref{tab:tab_hw_setting}).
Here, besides the computing constraint, we have additional limits on the physical line buffer size and thus use them to set upper bounds for model depth.

\section{Evaluation}
\label{sec:eval}

\subsection{Training and hardware settings}
\label{ssec:setting}

We use the same datasets and follow the same training settings in \cite{eCNN_2019}.
We also train conventional model structures for comparison.
We use 96-ch FFDNet as the denoising baseline; however, we remove BN layers and add a skip connection to improve training speed and stability.
The resultant model is called FFDNet*.
For the SR baseline, we directly use the 64-ch EDSR-baseline in \cite{EDSR_2017} which has the same model capability as SRResNet \cite{SRResNet_2017}.

The performance targets and computing capability are listed in Table \ref{tab:tab_hw_setting}.
We focus on high-performance Full-HD and 4K-UHD applications, and the computing capability is set to the same level as the eCNN processor in \cite{eCNN_2019}.
For each feature reusing target, we set an additional line buffer limit based on the line buffer size of the corresponding baseline model. In addition, we consider only the E3R1 variant for its smaller line buffer usage.

\subsection{Image quality}
\label{ssec:quality}

For denoising, we summarize the PSNR values of the polished models on test datasets in Table \ref{tab:tab_dn_psnr}.
We also include two reference numbers from the benchmark BM3D \cite{bm3d_2007} and the original FFDNet \cite{FFDNet_2018}.
We use twelve and five layers of FFDNet* as the baseline models for Full-HD and 4K-UHD throughputs respectively.
The E3R1 and E1R3 variants constantly show about 0.15-0.44 dB of PSNR gains over the baselines for all performance targets.
However, the E3R3 variant has significant quality drops because it has much fewer non-linear layers, in particular for the target A.

Similarly, we list the results for SR$\times$4 in Table \ref{tab:tab_srx4_psnr} and also include VDSR \cite{VDSR_2016} and SRResNet as reference numbers.
In this case, all ERModule variants show similar performance gains since they are all sufficiently deep for the training dataset.
In contrast, EDSR-baseline-B1 is up to 1.2 dB worse for the targets B and E due to its shallow five-layer depth.

\begin{table}[t]
\caption{PSNR (dB) of polished models for denoising.}
\begin{minipage}[b]{1.0\linewidth}
  \centering
  \centerline{\includegraphics[width=7.6cm]{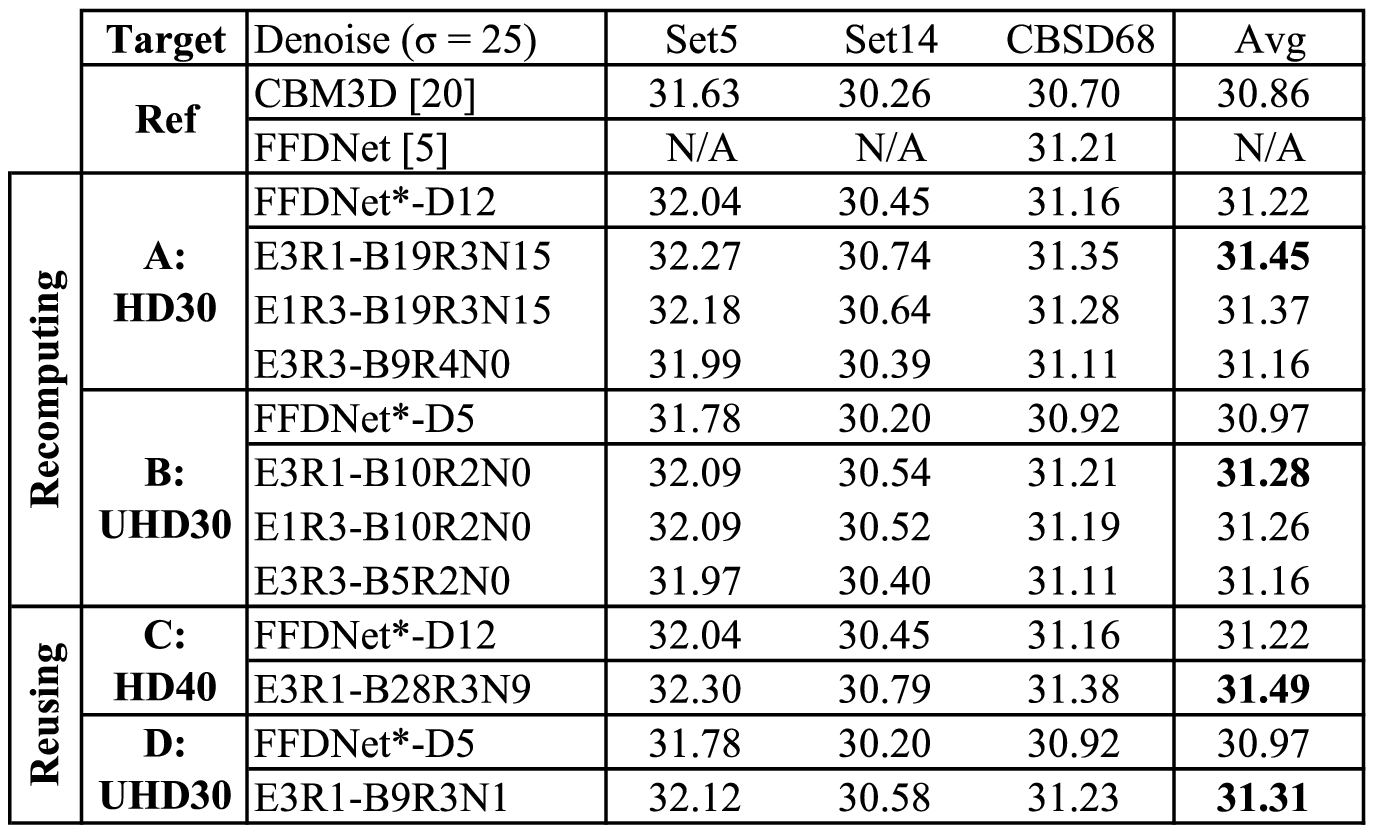}}
\end{minipage}
\vspace*{-0.8cm}
\label{tab:tab_dn_psnr}
\end{table}

\begin{table}[t]
\caption{PSNR (dB) of polished models for SR$\times$4.}
\begin{minipage}[b]{1.0\linewidth}
  \centering
  \centerline{\includegraphics[width=8.4cm]{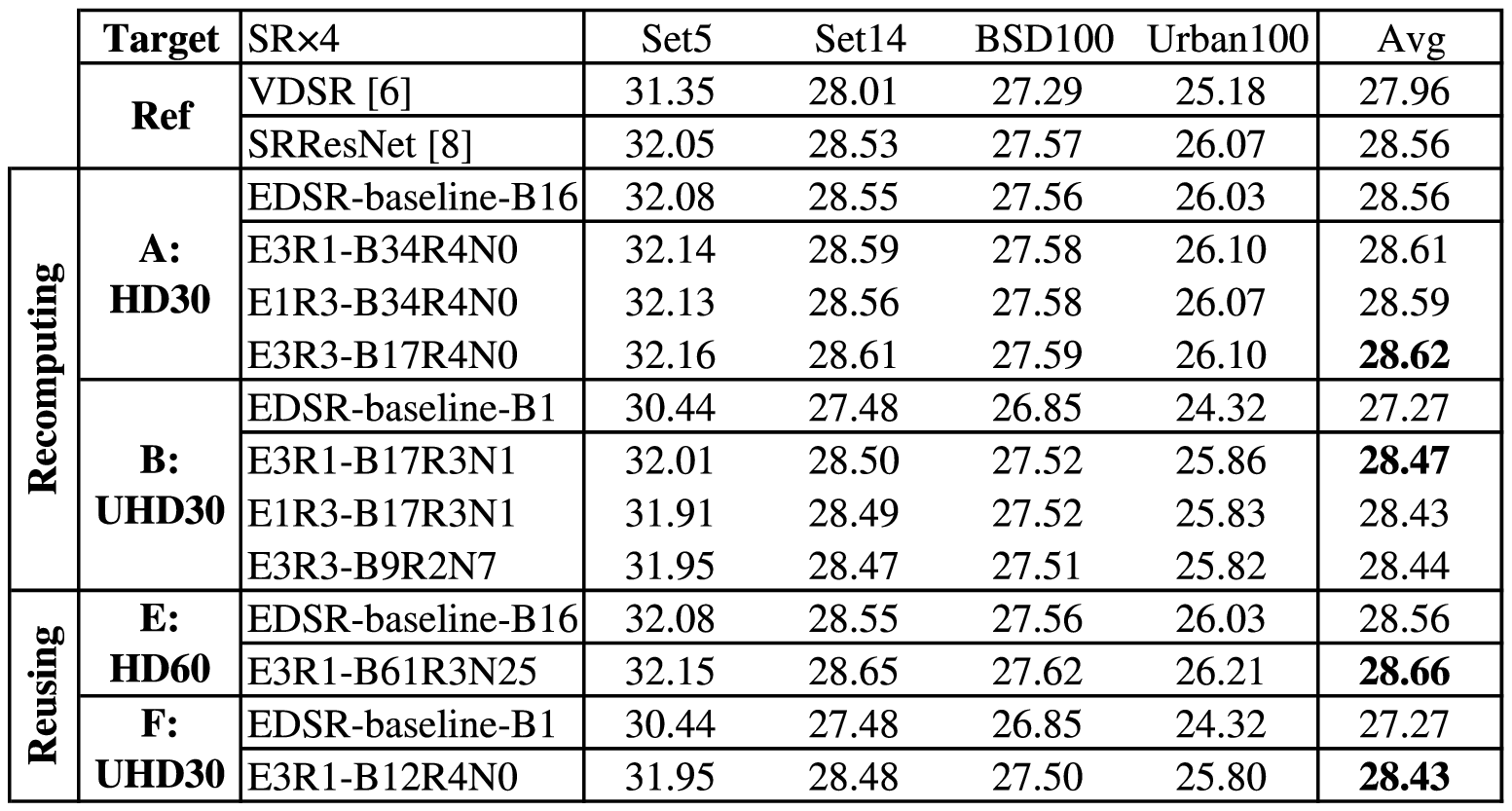}}
\end{minipage}
\vspace*{-0.8cm}
\label{tab:tab_srx4_psnr}
\end{table}

\subsection{Performance for feature recomputing}
\label{ssec:qual_recom}

\begin{figure}[t]
\begin{minipage}[b]{.48\linewidth}
  \centering
  \centerline{\includegraphics[width=4.0cm]{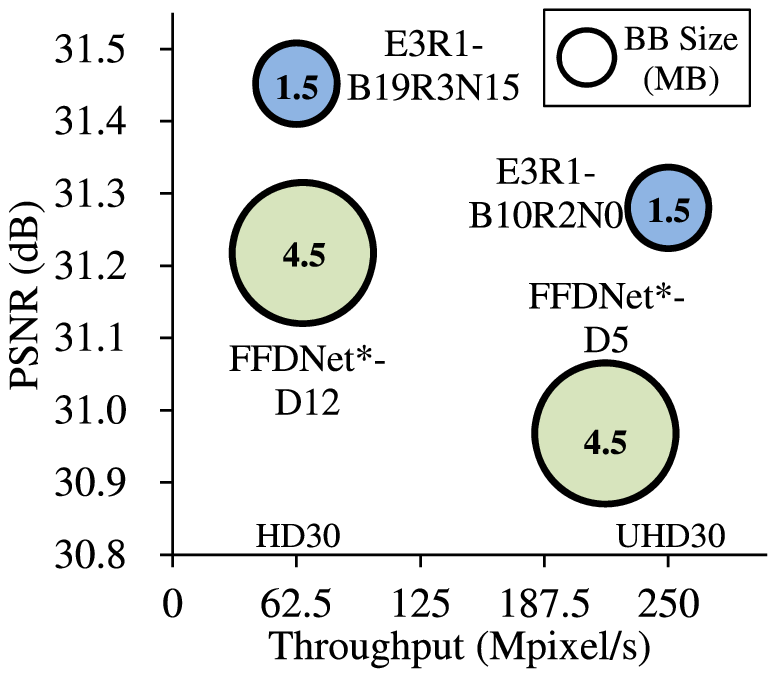}}
  \centerline{$\quad$(a) Denoising (DnERNet-12ch)}\medskip
\end{minipage}
\hfill
\begin{minipage}[b]{0.48\linewidth}
  \centering
  \centerline{\includegraphics[width=4.0cm]{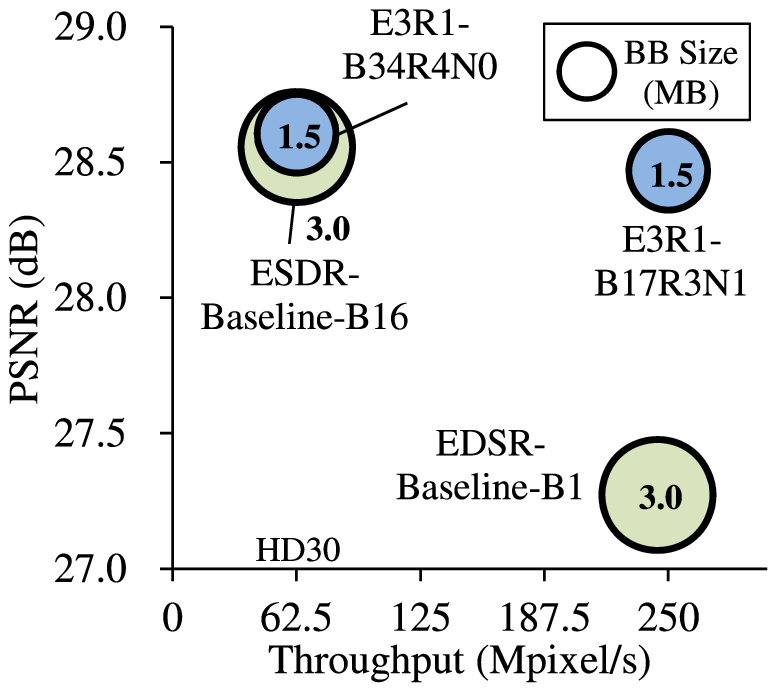}}
  \centerline{(b) SR$\times$4 (SR4ERNet)}\medskip
\end{minipage}
\vspace*{-0.6cm}
\caption{Performance with the feature recomputing flow for (a) denoising and (b) SR$\times$4 applications.
(BB: Block buffer)}
\label{fig:fig_qual_recom}
\end{figure}

We assume there are three block buffers deployed as the feature operands in \cite{eCNN_2019}.
Then the bubble chart in Fig. \ref{fig:fig_qual_recom} presents the performance benefits of the ERNets for the feature recomputing flow, where we only show the E3R1 variant for simplicity.
The bubble area represents the size of the deployed block buffers, and ERNets can use smaller block buffers for their fewer input/output channels in building blocks.
Therefore, ERNets provide better image quality and smaller block buffers than the conventional models while delivering the same or even higher pixel throughputs.

\subsection{Performance for feature reusing}
\label{ssec:qual_reuse}

\begin{figure}[t]
\begin{minipage}[b]{.48\linewidth}
  \centering
  \centerline{\includegraphics[width=4.0cm]{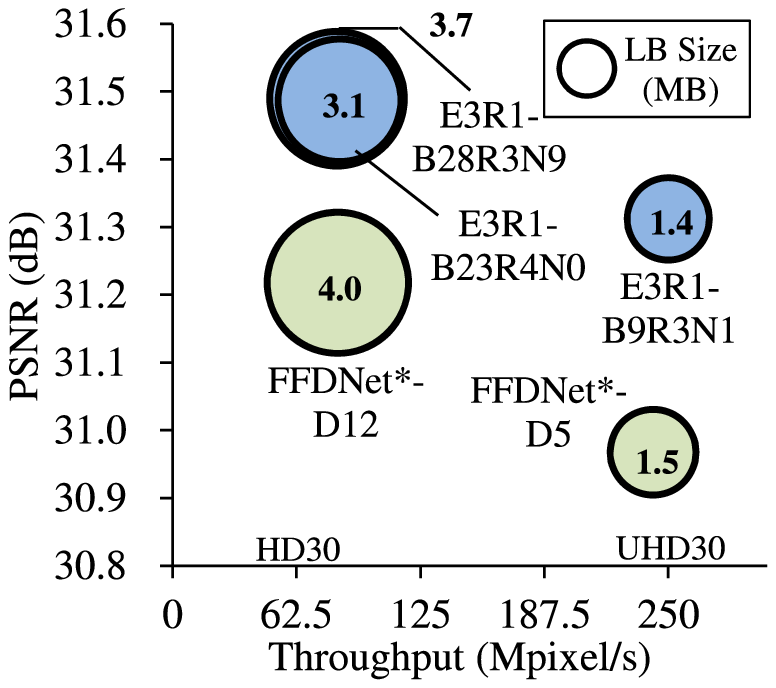}}
  \centerline{$\quad$(a) Denoising (DnERNet-12ch)}\medskip
\end{minipage}
\hfill
\begin{minipage}[b]{0.48\linewidth}
  \centering
  \centerline{\includegraphics[width=4.0cm]{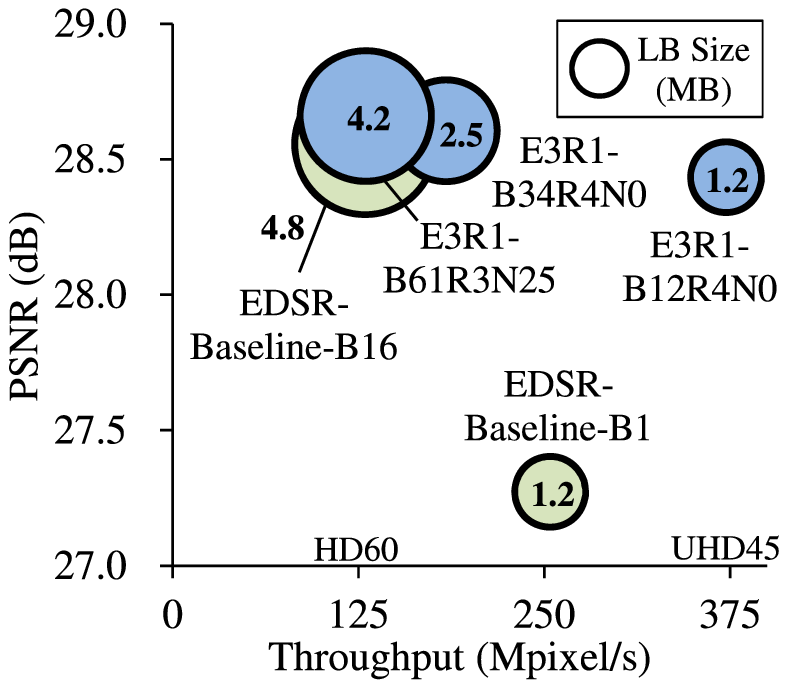}}
  \centerline{(b) SR$\times$4 (SR4ERNet)}\medskip
\end{minipage}
\vspace*{-0.6cm}
\caption{Performance with the feature reusing flow for (a) denoising and (b) SR$\times$4 applications.
(LB: Line buffer)}
\label{fig:fig_qual_reuse}
\end{figure}

For the feature reusing flow, we show the performance comparison in the bubble chart Fig. \ref{fig:fig_qual_reuse} and use the bubble area to represent the line buffer size since it is a major hardware cost.
Here we include two additional models, DnERNet-12ch-E3R1-B23R4N0 and SR4ERNet-E3R1-B34R4N0, to show that they can deliver similar image quality while using smaller line buffers than their counterparts, DnERNet-12ch-E3R1-B28R3N9 and SR4ERNet-E3R1-B61R3N25, in our optimization procedure.
From the results, we can draw a similar conclusion that ERNets provide better quality and smaller line buffers with similar or even higher throughputs.

\subsection{ReLU before residual connection}
\label{ssec:qual_relu}

In Fig. \ref{fig:fig_model}, we only apply ReLU in the middle of the expansion and reduction filters.
Alternatively, we can apply another ReLU before the residual connection to increase non-linearity.
We find that this alternative provides similar image quality in general and can compensate the cases of shallow model depth.
For example, it raises 0.12 dB for the five-layer E3R3-B5R2N0 for denoising in Table \ref{tab:tab_dn_psnr}.

\section{Conclusion}
\label{sec:conclusion}

In this paper, we discuss a novel model family---ERNet---to construct hardware-oriented CNN models.
It is devised for block-based inference flows which reduce excessive DRAM bandwidth by feature recomputing or reusing.
When building high-quality networks, it additionally considers temporary expansion-reduction layers which do not induce the hardware overheads for model deepening or widening.
According to the experiments for denoising and SR tasks, it can achieve better image quality and even higher pixel throughputs than the conventional model structures while using smaller block buffers and line buffers.
We also believe that ERNets can be extended to enhance image quality and hardware performance for more CNN tasks on resource-limited edge devices.




\end{document}